\documentclass[11pt]{article}

\usepackage[preprint]{acl}

\usepackage{tcolorbox}
\usepackage{comicneue}

\newtcolorbox{promptbox}[1][]{
  colback=gray!5,
  colframe=gray!40,
  boxrule=0.5pt,
  arc=4pt,
  left=8pt,
  right=8pt,
  top=8pt,
  bottom=8pt,
  fontupper=\small\comicneue,
  title=#1,
  coltitle=black,
  colbacktitle=gray!15,
  toptitle=4pt,
  bottomtitle=4pt,
  fonttitle=\bfseries\sffamily
}

\usepackage{times}
\usepackage{latexsym}
\usepackage{amsmath}
\usepackage{amssymb}
\usepackage{booktabs}
\usepackage{multirow}
\usepackage{array}
\usepackage[table]{xcolor}
\usepackage[T1]{fontenc}
\usepackage[utf8]{inputenc}
\usepackage{microtype}
\usepackage{inconsolata}
\usepackage{graphicx}

\title{X$^3$-OPD: Distilling Reasoning into Large Audio-Language Models via On-Policy Alignment}

\author{
  \textbf{Dongjie Fu\textsuperscript{1,2}},
  \textbf{Di Cao\textsuperscript{1}},
  \textbf{Xize Cheng\textsuperscript{2}},
  \textbf{Zihan Zhang\textsuperscript{2}},
  \textbf{Wenxu Jia\textsuperscript{2}},
  \\
  \textbf{Yifu Chen\textsuperscript{2}},
  \textbf{Shengpeng Ji\textsuperscript{1,2}},
  \textbf{Yu Zhang\textsuperscript{2}},
  \textbf{Tao Jin\textsuperscript{2}\thanks{Corresponding author.}}
  \\
  \\
  \textsuperscript{1}Tencent Hunyuan
  \qquad
  \textsuperscript{2}Zhejiang University
}

\begin{document}
\maketitle


\begin{abstract}

While audio large language models have achieved remarkable progress in auditory perception, they still lag behind text-based large language models in deep logical reasoning capabilities. This performance disparity primarily stems from the scarcity of high-quality audio reasoning data. To bridge this modality gap, we propose a novel cross-modal on-policy distillation framework. Our approach leverages a powerful text model as a teacher to provide dynamic correction and guidance to the audio foundational model during on-policy exploration. Furthermore, we construct a three-tier symmetric dataset encompassing logical reasoning, complex audio reasoning, and speech dialogue. By dynamically aligning the student model's reasoning trajectories within the high-noise auditory space with the teacher model's prior knowledge in the clean text space, our method effectively distills robust reasoning capabilities into the audio domain. Extensive experiments demonstrate that our model achieves substantial improvements on the audio and speech understanding benchmarks, paving a new path for comprehensive audio-language understanding.

\end{abstract}

\section{Introduction}


Recent advances in Large Audio-Language Models (LALMs) mark a significant shift from the traditional cascaded paradigm of pairing Automatic Speech Recognition with text-only Large Language Models (LLMs)\citep{zhang2023speechgpt,tang2023salmonn,xu2025qwen2}. At the perceptual level, LALMs natively model continuous acoustic signals and thereby preserve paralinguistic and environmental cues, including intonation, emotion, speaker turns, and sound events, that are irreversibly lost in intermediate transcription. More fundamentally, this native interface also reshapes the reasoning paradigm: instead of inferring over a sanitized textual transcript, the model can ground its inferences directly in the acoustic signal, leveraging prosodic emphasis, ambient context, and temporally structured sound events that have no faithful textual surrogate\citep{xu2025qwen3,tian2025stepaudior1technicalreport}. In doing so, LALMs open a path toward comprehensive audio-language understanding, in which perception and reasoning are unified end-to-end rather than fractured across a lossy transcription boundary.

Despite this perceptual advantage, a substantial reasoning gap persists between LALMs and text-only LLMs. The root cause is a fundamental data asymmetry: while modern LLMs are optimized over trillions of logically dense textual tokens\citep{openai2024gpt4technicalreport,guo2025deepseek}, audio foundation models are trained predominantly on shallow alignment objectives such as transcription, captioning, or short-form question answering. Eliciting multi-step reasoning over acoustic input therefore demands large-scale Audio Chain-of-Thought supervision, yet manually annotating step-wise logical trajectories over highly variable continuous audio is computationally intractable and prohibitively costly. A natural alternative is to transfer reasoning capability from a strong text teacher into an acoustic student via knowledge distillation, exploiting the abundance of textual logical supervision that LLMs already encode.

Early attempts in this direction adopt offline distillation, in which a frozen text teacher generates static reasoning trajectories that the student is trained to imitate\citep{xie2025audioreasonerimprovingreasoningcapability}. The supervision is then strictly tied to states the student never produces, and once the student's acoustic perception diverges from the teacher's textual premise, the offline target is out of distribution exactly where the student needs corrective signal the most. This cross-modal exposure bias is what motivates a recent line of on-policy distillation work in the audio--text setting~\citep{cao2026xopdcrossmodalonpolicydistillation,hu2026cordbridgingaudiotextreasoning}, which lets the student roll out under its own acoustic conditioning and aligns the resulting trajectories with the teacher through token-level KL-style supervision.

Despite these algorithmic advances, the cross-modal regime addressed by existing on-policy methods remains narrow. The teacher signal is sourced almost exclusively from text-only logical reasoning corpora, so the resulting student inherits chain-of-thought competence over textually transcribable content but remains under-supervised on two reasoning regimes that are constitutive of audio understanding: reasoning grounded in non-linguistic acoustic events whose evidence is not recoverable from a transcript, and reasoning conditioned on paralinguistic cues such as prosody, hesitation, and emotional shifts in spoken dialogue. Therefore, closing the audio-text reasoning gap demands a unified approach: advancing the distillation algorithm while fundamentally expanding the supervision distribution to encompass the regimes unique to the acoustic modality.

This work addresses both fronts jointly under a single framework, \textsc{X$^3$-OPD}, targeting three distinct tiers of audio reasoning. On the data side, a three-tier symmetric corpus is constructed---spanning textual reasoning rendered into speech, caption-grounded audio-event reasoning, and prosody-aware spoken dialogue---so that every instance admits both a text-side and an audio-side input under a uniform $(x_t, x_a, q, a^\star)$ schema. On the algorithmic side, a cross-modal on-policy distillation procedure is built on top of this corpus: the acoustic student rolls out reasoning trajectories conditioned on its own acoustic perception of $x_a$, while the text teacher scores those trajectories token-by-token under the matched $x_t$. Crucially, the teacher is granted privileged access to the ground-truth answer alongside $x_t$, allowing it to guide the student along a provably correct reasoning path. Furthermore, for open-ended queries, this token-level scoring allows the teacher to dynamically impart its own robust reasoning trace, rather than rigidly forcing the student to memorize a static target. This dynamic realignment occurs exactly at the steps where the student's perceptual premise drifts off the teacher's logical manifold. The two components are complementary: on-policy rollouts ensure the supervision lands at states the student actually visits, while the three-tier paired corpus ensures the teacher's supervision remains applicable across logical, acoustic, and paralinguistic reasoning alike.

\begin{itemize}
\item \textbf{A three-tier symmetric reasoning corpus} covering textual reasoning, audio-event reasoning, and paralinguistic spoken dialogue with paired text and audio inputs, extending cross-modal distillation supervision beyond the text-logical regime of prior on-policy methods.
\item \textbf{A cross-modal on-policy distillation framework} that scores trajectories from the student's acoustic-conditioned policy using the text teacher under matched text inputs, anchoring supervision at on-policy states while preserving robust text-conditioned signals.
\item \textbf{Comprehensive empirical evidence} on MMAU, MMSU, and BIG Bench Audio demonstrating substantial gains over strong offline and on-policy baselines, alongside text-domain capability preservation and robustness analyses isolating the contributions of cross-modal alignment.
\end{itemize}

\section{Related Work}
\subsection{Reasoning in Large Audio Language Models}
The transition from cascaded speech systems to LALMs has fundamentally transformed the modality interface, preserving rich acoustic and paralinguistic nuances within continuous representations \citep{ji2024wavchat,luo2026surveylargeaudiolanguage,fu-etal-2025-pachat}.
Early systems \citep{zhang2023speechgpt,huang2024audiogpt} couple discrete speech tokens or expert audio modules with an LLM, while end-to-end models \citep{chu2023qwen,chu2024qwen2,tang2023salmonn,kong2024audio} align continuous audio encoders with pretrained LLMs via lightweight adapters. More recent omni-modal systems \citep{xu2025qwen2,xu2025qwen3,gpt4o,comanici2025gemini} unify speech, audio, vision and text in a single backbone. 


The remarkable reasoning capabilities of Large Language Models \citep{yang2025qwen3,openai2024gpt4technicalreport} have catalyzed a paradigm shift in LALMs, extending the research focus from fundamental audio perception to complex reasoning. CoT prompting, which has demonstrated profound efficacy in textual reasoning, is increasingly recognized as equally pivotal for audio-centric tasks \citep{guo2026surveyaudioreasoningmultimodal}. Unlike purely text-based reasoning, audio reasoning necessitates extrapolating answers from continuous acoustic cues, such as speech emotion, prosody, and audio events. To address this, some existing methodologies construct audio-specific CoT datasets for Supervised Fine-Tuning (SFT) \citep{xie2025audioreasonerimprovingreasoningcapability,li2026audiocogitodeepaudioreasoning}. Moreover, Reinforcement Learning (RL) techniques have been recently introduced to encourage models \citep{tian2025stepaudior1technicalreport,zhong2025omnir1reinforcementlearningomnimodal}, which aim to cultivate genuine audio-driven reasoning, mitigating the model's over-reliance on textual priors.

However, the heavy reliance on SFT is bottlenecked by the prohibitive cost and complexity of audio CoT annotation, while biased data distributions further degrade generalization. Meanwhile, RL approaches are still hindered by sparse rewards \citep{fan2025incentivizingconsistenteffectivescalable}. Therefore, exploring new training paradigms for reasoning in LALMs remains an urgent necessity.

\subsection{Knowledge Distillation for Large Language Models}
Knowledge distillation (KD) \citep{hinton2015distillingknowledgeneuralnetwork} has traditionally served to transfer large teacher capabilities into compact students. Early text-domain extensions adopted offline forms, where students learned from static, teacher-generated corpora or reasoning traces \citep{kim2016sequencelevelknowledgedistillation,sanh2020distilbertdistilledversionbert}. Such supervision is strictly tied to sequences the student never produces, exposing the student to a train--inference distribution mismatch---commonly referred to as exposure bias---whose errors compound at test time. This motivates on-policy distillation, in which the student samples its own continuations and receives per-token teacher feedback at the states it actually visits. Generalized Knowledge Distillation (GKD) \citep{tan2023gkdgeneralknowledgedistillation} and MiniLLM \citep{gu2026minillmonpolicydistillationlarge} formalize this view and show that on-policy updates substantially mitigate exposure bias and provide a strictly stronger supervision signal once the inference policy drifts from any static training distribution. More recent work further unifies the dense token-level supervision of distillation with the on-policy nature of reinforcement learning, achieving substantial efficiency gains over reward-only RL while retaining its trajectory-level objective \citep{lu2025onpolicydistillation,song2026surveyonpolicydistillationlarge}.

The speech and audio community, by contrast, has historically employed distillation primarily for representation compression or self-distillation \citep{chang2022distilhubertspeechrepresentationlearning,baevski2022data2vecgeneralframeworkselfsupervised}, with limited engagement with generative reasoning. Although recent LALMs have extended distillation to speech-to-text generation \citep{milbich2020divadiversevisualfeature,li2026audiocogitodeepaudioreasoning}, they still rely on static, offline teacher traces and therefore inherit the exposure bias documented in the textual setting, which manifests as a pronounced reasoning gap between LALMs and their text-only counterparts. Very recent works begin to lift on-policy distillation beyond the unimodal text setting. X-OPD \citep{cao2026xopdcrossmodalonpolicydistillation} addresses the cross-modal case, aligning the two modalities through token-level KL on the student's on-policy rollouts. CORD \citep{hu2026cordbridgingaudiotextreasoning} applies an importance-weighted reverse-KL at the token level to prioritize critical early-step deviations. Both methods establish that on-policy KL-based supervision can effectively bridge the audio-text reasoning gap.

However, these advances are largely confined to distilling text-based logical reasoning, neglecting the fundamental essence of audio models: perception and reasoning grounded in continuous acoustic signals. When the student ingests an acoustic signal, the supervisory boundary becomes genuinely cross-modal, requiring the student's trajectories to be corrected under residual perceptual uncertainty, a regime that existing on-policy distillation methods fail to explicitly account. To bridge this critical gap, \textsc{X$^3$-OPD} is proposed to systematically extend the on-policy distillation paradigm into the cross-modal regime, aligning the student's reasoning trajectories directly within the continuous acoustic space.

\section{Preliminaries}

\subsection{Problem Setting}

This work targets reasoning distillation across the text--audio modality boundary. Let $\pi_T(\cdot \mid x_t, q, a^\star; \theta_T)$ denote a frozen text teacher with parameters $\theta_T$, taking a text input $x_t$, a question $q$ and ground truth answer $a^\star$ and let $\pi_S(\cdot \mid x_a, q; \theta_S)$ denote a trainable audio student with parameters $\theta_S$, taking an acoustic input $x_a$ and the same question. Throughout this work, a symmetric sample is a tuple $(x_t, x_a, q, a^\star)$ in which $x_t$ and $x_a$ encode the same underlying content in their respective modalities, and $a^\star$ is the ground truth. A reasoning trajectory $y = (y_1, \dots, y_L)$ is the concatenation of an intermediate chain-of-thought and a final answer, sampled autoregressively from either $\pi_T$ or $\pi_S$.

\begin{figure*}[t]
  \centering
  \includegraphics[width=\textwidth]{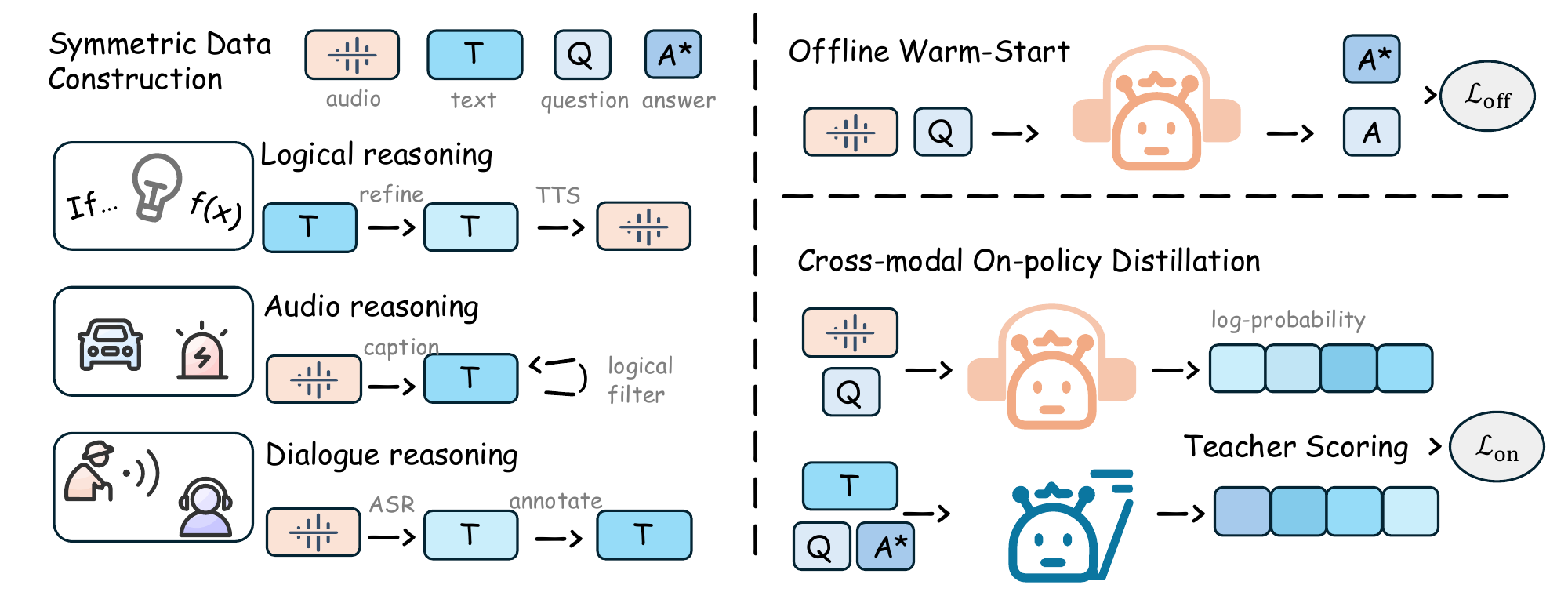}
  \caption{
  Overview of \textsc{X$^3$-OPD}.
  \textbf{Left:} a three-tier symmetric corpus---Logical (text + TTS speech),
  Audio (audio + refined caption), and Dialogue (dialogue + prosody-aware
  meta-caption)---providing $(x_t, x_a, q, a^\star)$ for every instance.
  \textbf{Right:} an offline warm-start ($\mathcal{L}_{\mathrm{off}}$,
  Eq.~\ref{eq:offline}) followed by cross-modal on-policy distillation,
  where the student rolls out under $(x_a, q)$ and the frozen teacher
  scores the same trajectories under $(x_t, q, a^\star)$ to compute
  $\mathcal{L}_{\mathrm{on}}$ (Eq.~\ref{eq:objective}).
}

  \label{fig:framework}
\end{figure*}

\subsection{Offline Distillation and its Limitation}

The standard offline distillation objective trains the student to imitate teacher trajectories sampled once and held fixed:

\begin{equation} 
\begin{aligned} 
\mathcal{L}_{\mathrm{off}}(\theta_S) \;&=\; \mathbb{E}_{(x_t, x_a, q) \sim \mathcal{D}} \\ 
&\quad \, \mathbb{E}_{y \sim \pi_T(\cdot \mid x_t, q)} \!\left[ -\log \pi_S(y \mid x_a, q; \theta_S) \right]. 
\end{aligned} 
\label{eq:offline} 
\end{equation}
In the unimodal text setting, where $x_t = x_a$ and $\pi_T$, $\pi_S$ share an input space, Eq.~\eqref{eq:offline} is a faithful proxy for the population KL between teacher and student. Once the input modalities diverge, two issues emerge with this offline objective that neither standard unimodal distillation nor prior on-policy work in text~\citep{lu2025onpolicydistillation} is equipped to handle.

\paragraph{Perception–text divergence} The student's input $x_a$ is an acoustic signal whose semantic content is recovered only up to a posterior $p(s \mid x_a)$ over latent percepts $s$. The teacher reasons over $x_t$, which is by construction a single point in this posterior; its trajectories therefore implicitly assume a perceptual premise that the student rarely matches token-for-token at training time and almost never matches under noise or accent shift at test time.

\paragraph{Exposure bias across modalities} Because the teacher trajectories in Eq.~\eqref{eq:offline} are independent of $\theta_S$, the student is supervised only at states it would not have visited on its own. As soon as the student commits to a different intermediate token, which frequently occurs when its acoustic perception diverges from $x_t$, the offline target is out of distribution with respect to the student's induced state distribution, a cross-modal instance of the exposure bias studied in sequence modeling~\citep{bengio2015scheduledsamplingsequenceprediction}. The supervisory signal is then strongest exactly where it is least informative.

A direct consequence of these two failure modes is that $\mathcal{L}_{\mathrm{off}}$ cannot teach the student how to reason given that its own perception may be wrong. Section~\ref{sec:method} addresses both by sampling trajectories from $\pi_S$ itself and scoring them with $\pi_T$ on the matched text input.

\section{Method} \label{sec:method}

The proposed framework has three components: (i) a three-tier modality-paired corpus that establishes a uniform $(x_t, x_a, q, a^\star)$ schema across textual reasoning, audio-event reasoning, and spoken dialogue; (ii) a cross-modal on-policy distillation objective that scores student rollouts under the matched text input; and (iii) a training recipe that combines a brief offline warm-start with on-policy optimization. An overview is given in Figure~\ref{fig:framework}.

\subsection{Three-Tier Symmetric Data Construction} \label{sec:data}To comprehensively enhance the model's reasoning capabilities in multimodal environments, we construct a dataset encompassing three distinct levels of difficulty and modal emphases. Specifically, this dataset defines a three-tier hierarchy of reasoning tasks: pure logical reasoning based on textual semantic transcription, audio event reasoning grounded in complex acoustic scenes, and spoken dialogue interaction reasoning contingent on paralinguistic and social cues. Through this three-tier data construction, we aim to guide the model to progressively transition from fundamental semantic and logical understanding to the deep parsing of real-world physical acoustic scenes.
\paragraph{Tier 1: Textual reasoning rendered into speech.} Large-scale text-only reasoning corpora \cite{lambert2025tulu,shao2023characterllmtrainableagentroleplaying} provide the textual side $x_t$, the question $q$, and the gold answer $a^\star$. To ensure the naturalness of the data, the generation of the acoustic side $x_a$ undergoes a rigorous pipeline: first, the original text corpora are rewritten into a format more suitable for spoken expression. Subsequently, they are rendered using a high-fidelity text-to-speech (TTS) system and subjected to closed-loop validation via an automatic speech recognition (ASR) system to filter out audio data with poor generation quality. This tier supplies dense logical supervision and serves as the primary foundational source for the model's reasoning signals.
\paragraph{Tier 2: Audio reasoning grounded in captions.} Tier 2 samples real audio clips paired with structured captions that describe salient acoustic events, sound sources, and temporal structures. These captions initially derive from existing open-source datasets \citep{drossos2019clothoaudiocaptioningdataset,kim-etal-2019-audiocaps} and are subsequently reconstructed using advanced captioning models \citep{xu2025qwen3}, followed by rigorous human verification to ensure the precision and dimensional richness of the acoustic feature descriptions. On this basis, we strictly filter the data according to the logical consistency among the acoustic descriptions, questions, and answers, removing samples with unreasonable reasoning chains or factual errors. The teacher model is queried with these high-quality captions as $x_t$ to generate a chain-of-thought that explicitly cites acoustic evidence at each reasoning step; conversely, the student model directly receives the raw audio as $x_a$ alongside the same question $q$. 
\paragraph{Tier 3: Spoken dialogue with prosody-aware captions.} 
Tier 3 covers reasoning that hinges on paralinguistic cues---such as turn boundaries, hesitations, emphases, tonal shifts, and subtle emotional fluctuations---complex semantic dimensions that ordinary plain-text transcripts fail to convey. Each instance pairs a multi-turn spoken dialogue from a real-world scenario with a meta-caption, which deeply augments the transcript with precise prosodic and turn-level annotations. The teacher model processes the meta-caption as $x_t$, enabling it to capture the underlying subtext behind the dialogue; the student model learns directly from the raw speech waveform as $x_a$. The design of this tier requires the model not only to understand what is said but also to comprehend how it is said, significantly enhancing the model's empathic capacity and its acuity in recognizing dialogue intentions within complex human-computer interaction scenarios.

During the actual training process, data from these three distinct tiers are fully shuffled and mixed. This joint optimization strategy allows the student model to seamlessly integrate textual logical reasoning, audio feature parsing, and spoken dialogue interaction capabilities under a single training objective, effectively facilitating knowledge transfer across different modal features.

\subsection{Offline Warm-start} 
On-policy optimization is unstable when the student's initial rollouts are far from the teacher's distribution. To stabilize the early phase, $\pi_S$ is first fine-tuned for a short SFT pass on Eq.~\eqref{eq:offline} using teacher-generated trajectories. The warm-start is deliberately short, to seat the student inside the teacher's reasoning style, but not long enough to entrench the offline failure modes analyzed in Section~3.

\subsection{Cross-modal On-policy Distillation} \label{sec:opd}

\paragraph{On-policy Multi-path Rollout.}
For each symmetric sample, the student model first performs autoregressive generation under its current policy $\pi_S(\theta_S)$, conditioned on its own acoustic perception $x_a$ and the question $q$. To mitigate the high variance inherent in gradient estimation from single-trajectory rollouts and to explore the policy space more extensively, we independently sample $K$ candidate trajectories $Y=\{y^{(1)}, y^{(2)}, \dots, y^{(K)}\}$ for each input. Because these trajectories are drawn directly from the student model, they accurately reflect the actual state space visited under the current acoustic premise, ensuring that the subsequent supervisory signals strictly adhere to on-policy conditions.

\paragraph{Cross-Modal Teacher Scoring.}
To ensure that the student model $\pi_S$ faithfully inherits the logical reasoning capabilities of the teacher model $\pi_T$, we introduce a cross-modal advantage function. For the $k$-th sampled trajectory $y^{(k)}$, let $y_t^{(k)}$ be the $t$-th token. The highly capable teacher model evaluates this exact trajectory token-by-token, conditioned on the matched text input $x_t$ together with the gold answer $a^\star$, so that its token-level log-probabilities form a high-confidence reference distribution anchored to the correct conclusion.
The cross-modal advantage $A(y_t^{(k)})$ is defined to bridge the log-probability gap between the teacher's text-driven logical distribution and the student's speech-conditioned output:
\begin{equation}
\begin{split}
A(y_t^{(k)}) &= \log\pi_T(y_t^{(k)} \mid x_t, q, a^*,  y_{<t}^{(k)}) \\
&\quad - \log\pi_S(y_t^{(k)} \mid x_a, q, y_{<t}^{(k)})
\end{split}
\end{equation}
By calculating this advantage, the text-driven teacher model provides real-time cross-modal guidance signals at the exact states where the student's acoustic representations push it off the correct logical manifold.

\paragraph{Cross-Modal Distillation Objective.}
Integrating the multi-path rollout and the cross-modal advantage, we formulate a policy gradient-based distillation loss. Utilizing the $K$ sampled trajectories, the optimization objective is formulated to maximize the expected cross-modal advantage, achieving cross-modal reasoning transfer:
\begin{equation}
\begin{split}
\mathcal{L}_{\mathrm{on}}(\theta_S) &= -\mathbb{E}_{(x_t, x_a, q)}\Bigg[\frac{1}{K}\sum_{k=1}^K\frac{1}{|y^{(k)}|} \\
&\quad \sum_{t=1}^{|y^{(k)}|}r_{k,t}(\theta_S)A(y_t^{(k)})\Bigg]
\end{split}
\label{eq:objective} 
\end{equation}
where $r_{k,t}(\theta_S)=\frac{\pi_S(y_t^{(k)} \mid x_a, q, y_{<t}^{(k)}; \theta_S)}{\pi_{old}(y_t^{(k)} \mid x_a, q, y_{<t}^{(k)})}$ represents the probability ratio between the current policy $\pi_S$ and the sampling policy $\pi_{old}$. This formulation yields a single, highly efficient objective: it is strictly on-policy and suited for open-ended instructional data lacking gold standard answers, enabling the student model to internalize the teacher's open-ended reasoning capabilities purely through cross-modal alignment.

\section{Experiments}

\subsection{Experimental Setup}

\paragraph{Backbones.}
The student is initialized from Qwen3-Omni-30B-A3B-Thinking ~\citep{xu2025qwen3}, whose acoustic encoder and LLM backbone are kept jointly trainable while the speech tokenizer is frozen. The teacher is Qwen3-235B-A22B-Thinking ~\citep{yang2025qwen3}, held frozen throughout. The teacher reads $x_t$ and the student reads $x_a$ under the modality-paired schema of Section~\ref{sec:data}.

\paragraph{Training data.}
The three-tier corpus contains 27.8K Tier-1, 32K Tier-2, and 12K Tier-3 instances, fully shuffled at the batch level. Tier-1 takes text prompts from Tulu~3~\citep{lambert2025tulu} and NaturalReasoning~\citep{yuan2025naturalreasoning}, rewrites them into spoken style with Gemini-3-Flash-Preview, synthesizes audio via CosyVoice~3~\citep{du2025cosyvoice3} at $24$\,kHz, and filters with SenseVoice~\citep{speechteam2024funaudiollm} ASR back-translation at WER$\!\le\!5\%$. Tier-2 draws audio and captions from AudioCaps~\citep{kim-etal-2019-audiocaps} and Clotho~v2.1~\citep{drossos2019clothoaudiocaptioningdataset}, with captions reconstructed by Qwen3-Omni-30B-A3B-Captioner and human-verified on a $5\%$ subset before caption--question--answer consistency filtering. Tier-3 is built on dyadic and multi-party spoken dialogues from IEMOCAP~\citep{busso2008iemocap} and MELD~\citep{poria2019meldmultimodalmultipartydataset}, augmented with prosody- and turn-level meta-captions. 
The offline warm-start SFT pass uses a 12K subset sampled under a tier ratio of $5\!:\!3\!:\!2$, deliberately over-weighting Tier-1 and Tier-3 to anchor the model's logical reasoning format and sensitivity to paralinguistic cues before the RL stage.

\paragraph{Training configuration.}
Offline warm-start: $1$ epoch, batch size $128$, learning rate $1\!\times\!10^{-5}$, cosine schedule. Cross-modal on-policy stage: $K\!=\!4$ rollouts per sample, AdamW with constant learning rate $2\!\times\!10^{-6}$. Rollouts use \texttt{vLLM}~\citep{kwon2023efficientmemorymanagementlarge} at temperature $1.0$. All runs are in the verl~\citep{sheng2025hybridflow} framework on $32\!\times\!$ NVIDIA H20.

For a fair comparison, we additionally train two baselines, all initialized from the same Qwen3-Omni-30B-A3B-Thinking student checkpoint:
(1) Standard SFT, fine-tuned on a supervised corpus in which each prompt is paired with a reasoning trace produced offline by the text-only teacher Qwen3-235B-A22B-Thinking together with the ground-truth answer, so that the student learns to imitate the teacher's chain-of-thought and to emit the correct final response;
(2) Generalized Knowledge Distillation (GKD)~\citep{tan2023gkdgeneralknowledgedistillation},
which optimizes the student policy against the same teacher under the forward KL divergence on on-policy student rollouts.

\begin{table*}[t]
\centering
\small
\setlength{\tabcolsep}{6pt}
\renewcommand{\arraystretch}{1.1}
\resizebox{\textwidth}{!}{%
\begin{tabular}{l cccc c ccc c c}
\toprule
& \multicolumn{4}{c}{\textbf{MMSU}} & & \multicolumn{3}{c}{\textbf{MMAU}} & & \textbf{BIG Bench} \\
\cmidrule(lr){2-5}\cmidrule(lr){7-9}
\textbf{Model} & Sem. & Phon. & Style & Traits & & Sound & Speech & Music & & \textbf{Audio} \\
\midrule
\rowcolor{gray!15}\multicolumn{11}{l}{\emph{Closed-Source LALMs}} \\
GPT-4o-Audio\citep{gpt4o}        & 59.7 & 41.6 & 21.4 & 39.7          & & 64.6 & 66.7 & 56.3          & & 88.5 \\
Gemini-3-Flash\citep{comanici2025gemini}      & 70.2 & 53.6 & 38.5 & \textbf{46.1} & & 75.4 & 80.6 & 71.0          & & 80.8 \\
\midrule
\rowcolor{gray!15}\multicolumn{11}{l}{\emph{Open-Weight LALMs}} \\
Qwen2-Audio-Instruct\citep{xu2025qwen2} & 38.5 & 29.4 & 21.7 & 26.1 & & 55.0 & 42.0 & 51.0           & & 32.5 \\
GLM-4-Voice\citep{zeng2024glm}          & 41.2 & 31.8 & 24.5 & 28.4 & & 52.4 & 50.6 & 49.1           & & 35.2 \\
Qwen2.5-Omni-7B\citep{xu2025qwen25omni}      & 55.1 & 37.3 & 39.4 & 42.5 & & 67.9 & 59.8 & 69.2           & & 62.2 \\
Qwen3-Omni-Instruct\citep{xu2025qwen3}  & 81.6 & 70.0 & 54.0 & 35.3 & & 80.5 & 78.1 & 74.3           & & 85.7 \\
Qwen3-Omni-Thinking\citep{xu2025qwen3}  & 81.8 & \textbf{73.2} & \textbf{65.2} & 42.7 & & 85.7 & 81.5 & \textbf{74.9} & & 87.9 \\
\midrule
\rowcolor{gray!15}\multicolumn{11}{l}{\emph{Ablation Baselines}} \\
SFT ($\mathcal{L}_{\mathrm{off}}$) & 82.1 & 69.7 & 62.4 & 42.8 & & 86.4 & 81.9 & 70.7 & & 86.7 \\
GKD                       & 80.2 & 68.4 & 63.1 & 41.1 & & 85.1 & 82.3 & 69.2 & & 84.3 \\
\midrule
\textbf{\textsc{X$^3$-OPD} (ours)}
  & \textbf{83.7} & 72.1 & 64.2 & 42.9
  & & \textbf{88.3} & \textbf{82.9} & 72.4
  & & \textbf{93.6} \\
\bottomrule
\end{tabular}%
}
\caption{Main evaluation results on MMSU, MMAU, and BIG Bench Audio. MMSU evaluation dimensions include Semantics (Sem.), Phonology (Phon.), Style, and Traits. \textbf{Bold} indicates the best performance in each column.}
\label{tab:main}
\end{table*}

\paragraph{Evaluation benchmarks.}
We evaluate our framework across three levels of domain shift. The in-domain evaluation covers benchmarks that directly probe audio-grounded reasoning: MMSU~\citep{wang2026mmsumassivemultitaskspoken} along its four axes (Semantics, Phonology, Style, Traits), MMAU~\citep{sakshi2024mmaumassivemultitaskaudio} on Sound/Speech/Music, and BIG Bench Audio aggregating acoustic QA, multi-event reasoning, audio-temporal grounding, and paralinguistic inference. We then stress-test intermediate reasoning quality on the held-out MMAR benchmark, where we evaluate two CoT-quality metrics, \emph{Rubrics} and \emph{CRS}. Finally, we evaluate out-of-distribution generalization on benchmarks with significant domain shifts, including the audio-visual WorldSense~\citep{xu2025qwen3}, video reasoning DailyOmni.

\subsection{Main Results}
\label{sec:main}

Table~\ref{tab:main} contrasts \textsc{X$^3$-OPD} with state-of-the-art closed- and open-source systems, as well as controlled in-house baselines. We summarize our core observations below.

\paragraph{Significant Enhancements in Cross-Modal Reasoning.}
Built upon the Qwen3-Omni-Thinking baseline, \textsc{X$^3$-OPD} demonstrates substantial improvements in complex audio-grounded reasoning. Compared to its base model, \textsc{X$^3$-OPD} not only pushes the BIG Bench Audio (BAB) score from $87.9$ to $93.6$ (a $+5.7$ improvement) but also achieves solid gains in semantic and acoustic event understanding (e.g., $+1.9$ on MMSU-Semantics and $+2.6$ on MMAU-Sound). Furthermore, \textsc{X$^3$-OPD} exhibits highly competitive performance against proprietary systems, surpassing GPT-4o-Audio on BAB by $+5.1$ points.

\paragraph{Mitigating Perceptual Regression via On-Policy Distillation.}
A comparison with off-policy methods reveals a structural exposure bias in static alignment. While off-policy SFT slightly improves high-signal semantic tasks over the Qwen3-Omni-Thinking baseline, it actively regresses on low-signal perceptual axes (Phonology $-3.5$, Style $-2.8$, Music $-4.2$). \textsc{X$^3$-OPD} effectively mitigates this degradation. By allowing the student to receive teacher feedback on its own trajectory, on-policy distillation substantially reduces these performance drops on perceptual tasks while achieving remarkable gains on reasoning-heavy axes like BAB.

\paragraph{Residual Gaps on Perceptual Axes.}
Despite the overall improvements, \textsc{X$^3$-OPD} still slightly trails the base model on three purely perceptual axes: MMSU-Phonology, MMSU-Style, and MMAU-Music. We attribute this to the intrinsic properties of cross-modal distillation: the reward signals provided by the text-side teacher propagate predominantly along semantic and logical pathways, leaving non-semantic acoustic representations (such as music and pure paralinguistic style) with insufficient gradient updates during optimization. Remarkably, even though the model is trained on synthesized data streams specifically designed to elicit logical reasoning---presenting an inevitable distribution shift---the performance degradation on these pure perception benchmarks remains exceptionally minimal (drop $\leq 2.6$).

\subsection{CoT-quality Evaluation}
\label{sec:mmar}

To rigorously assess the intermediate reasoning quality of the model on multi-step audio tasks, we evaluate \textsc{X$^3$-OPD} on the MMAR benchmark~\citep{ma2025mmar}. Our evaluation specifically focuses on the unambiguous Sound and Speech splits, with a primary emphasis on assessing the quality of the CoT itself. Beyond standard accuracy, we adopt the Rubrics and CRS (Correct Reasoning Score) metrics~\citep{ma2026interspeech2026audioreasoning} to quantify whether the generated reasoning steps are well-grounded in the acoustic premise. As shown in Table~\ref{tab:mmar}, \textsc{X$^3$-OPD} steadily improves accuracy over the Qwen3-Omni-Thinking baseline (Sound $+1.6$, Speech $+1.3$) while simultaneously elevating the CoT quality scores (Rubrics $+2.4$, CRS $+0.01$). This concurrent gain in both correct predictions and reasoning structure confirms that on-policy distillation genuinely anchors intermediate steps to the acoustic input, rather than merely exploiting statistical shortcuts to reach a final answer. Furthermore, compared to contemporary baselines like Step-Audio-R1---which suffers a severe performance collapse on the non-speech Sound split ($32.5\%$) due to speech-biased reward signals---\textsc{X$^3$-OPD} maintains robust and balanced audio reasoning capabilities, demonstrating that it effectively enhances \emph{how} to reason without relying on any MMAR-specific domain supervision.
\begin{table}[t]
\centering
\renewcommand{\arraystretch}{1.1}
\resizebox{\columnwidth}{!}{
\begin{tabular}{l cc c cc}
\toprule
& \multicolumn{2}{c}{\textbf{MMAR Acc. (\%)}} & & \multicolumn{2}{c}{\textbf{CoT Quality}} \\
\cmidrule(lr){2-3}\cmidrule(lr){5-6}
\textbf{Model} & Sound & Speech & & Rubrics & CRS \\
\midrule
Step-Audio-R1         & 32.5 & 68.7 & & 46.6 & 0.79 \\
Audio-Reasoner           & 42.4 & 42.5 & & 28.4 & 0.68 \\
Qwen3-Omni-Thinking      & 64.2 & 79.3 & & 58.0 & 0.85 \\
\midrule
\textbf{\textsc{X$^3$-OPD} (ours)}
                         & \textbf{65.8} & \textbf{80.6} & & \textbf{60.4} & \textbf{0.86} \\
\bottomrule
\end{tabular}%
}
\caption{MMAR results and CoT-quality scores.
\textbf{Bold} marks the best result in each column.}
\label{tab:mmar}
\end{table}

\subsection{Capability Preservation under Domain Shift}
\label{sec:cross}

A critical concern when fine-tuning on modality-specific data is the catastrophic forgetting of unobserved domains. Table~\ref{tab:cross} stress-tests the models on two out-of-distribution benchmarks absent from our corpus: audio-visual reasoning (WorldSense) and video reasoning (DailyOmni). We report both the raw score and the absolute performance drop ($\Delta$) relative to the base model.

The results demonstrate that off-policy imitation methods suffer from severe cross-modal degradation. Both SFT and GKD lose substantial ground across visual and video tasks, dropping by $5.3$ to $6.6$ points. This indicates that static token-level imitation of a text teacher induces a global policy drift within the shared LLM backbone. In contrast, \textsc{X$^3$-OPD} preserves the base model's initialization almost perfectly, with marginal variations ($-1.2$ and $-2.3$). By dynamically scoring the student's on-policy rollouts using the cross-modal advantage (Eq.~\ref{eq:objective}), \textsc{X$^3$-OPD} inherently minimizes the reverse-KL divergence against the highly capable, frozen text teacher. Because this teacher maintains strong general-purpose reasoning capabilities across unobserved domains, its dense token-level feedback acts as a continuous capability anchor, effectively restricting the student from drifting into degenerate, modality-specific shortcuts. This confirms that advanced audio reasoning can be cultivated without paying a multimodal tax on pre-existing vision or video capabilities.

\begin{table}[t]
\centering
\renewcommand{\arraystretch}{1.1}
\resizebox{\columnwidth}{!}{%
\begin{tabular}{l cc cc}
\toprule
& \multicolumn{2}{c}{\textbf{WorldSense}} & \multicolumn{2}{c}{\textbf{DailyOmni}} \\
\cmidrule(lr){2-3}\cmidrule(lr){4-5}
\textbf{Model} & Score & $\Delta$ & Score & $\Delta$ \\
\midrule
Qwen3-Omni-Thinking                & 54.0 & ---            & 75.8 & ---            \\
\midrule
SFT ($\mathcal{L}_{\mathrm{off}}$)    & 48.7 & $-5.3$        & 69.2 & $-6.6$        \\
GKD                           & 47.9 & $-6.1$        & 69.5 & $-6.3$        \\
\midrule
\textbf{\textsc{X$^3$-OPD} (ours)} & \textbf{52.8} & $\mathbf{-1.2}$ & \textbf{73.5} & $\mathbf{-2.3}$ \\
\bottomrule
\end{tabular}%
}
\caption{Capability preservation under domain shift. WorldSense (audio-visual) and DailyOmni (video) represent modalities completely absent from our fine-tuning corpus.}
\label{tab:cross}
\end{table}

\section{Conclusion}

In this work, we introduced \textsc{X$^3$-OPD}, a novel cross-modal on-policy distillation framework designed to bridge the reasoning gap between text-based Large Language Models and Large Audio-Language Models (LALMs). To address the exposure bias and perception-text divergence inherent in standard offline distillation, our approach dynamically aligns the student's acoustic-conditioned reasoning trajectories with the teacher's text-grounded prior knowledge. Supported by a newly constructed three-tier symmetric corpus---spanning textual reasoning, audio-event understanding, and paralinguistic spoken dialogue---\textsc{X$^3$-OPD} provides comprehensive supervision across multiple acoustic reasoning regimes. Extensive experiments demonstrate that our framework achieves substantial gains on major audio-reasoning benchmarks while mitigating catastrophic forgetting, ultimately establishing a robust paradigm for cross-modal reasoning directly from continuous acoustic signals.

\section*{Limitations}

While \textsc{X$^3$-OPD} significantly advances audio-grounded reasoning, several limitations remain to be addressed in future work. First, our current three-tier symmetric corpus does not yet encompass the full spectrum of acoustic phenomena. Specifically, complex musical reasoning and highly specialized, non-linguistic soundscapes are underrepresented in the training data, which is reflected in the residual performance gaps on purely perceptual axes like MMAU-Music. Expanding the dataset construction pipeline to encompass a broader diversity of audio domains is a critical next step. 

Second, the current framework relies exclusively on a text-based teacher's token-level log-probabilities as the supervisory signal. Because the text teacher is inherently blind to non-semantic acoustic cues that cannot be perfectly captured by captions or transcripts, this reward signal is somewhat constrained. It tends to bias the learning process toward semantic logic while under-penalizing perceptual errors in pure acoustic or paralinguistic contexts. To overcome this limitation, future work must introduce more multi-dimensional reward mechanisms. Incorporating direct audio-based reward models (RM), environmental verifiers, or rule-based multi-objective signals will be vital for cultivating a truly comprehensive and autonomous reasoning capability in LALMs.

\bibliography{custom}

@article{gpt4o,
  title={GPT-4o System Card},
  author={OpenAI},
  journal={
https://cdn.openai.com/gpt-4o-system-card.pdf},
  year={2024}
}

@inproceedings{huang2024audiogpt,
  title={Audiogpt: Understanding and generating speech, music, sound, and talking head},
  author={Huang, Rongjie and Li, Mingze and Yang, Dongchao and Shi, Jiatong and Chang, Xuankai and Ye, Zhenhui and Wu, Yuning and Hong, Zhiqing and Huang, Jiawei and Liu, Jinglin and others},
  booktitle={Proceedings of the AAAI Conference on Artificial Intelligence},
  volume={38},
  number={21},
  pages={23802--23804},
  year={2024}
}

@article{kong2024audio,
  title={Audio flamingo: A novel audio language model with few-shot learning and dialogue abilities},
  author={Kong, Zhifeng and Goel, Arushi and Badlani, Rohan and Ping, Wei and Valle, Rafael and Catanzaro, Bryan},
  journal={arXiv preprint arXiv:2402.01831},
  year={2024}
}

@article{zhang2023speechgpt,
  title={Speechgpt: Empowering large language models with intrinsic cross-modal conversational abilities},
  author={Zhang, Dong and Li, Shimin and Zhang, Xin and Zhan, Jun and Wang, Pengyu and Zhou, Yaqian and Qiu, Xipeng},
  journal={arXiv preprint arXiv:2305.11000},
  year={2023}
}

@article{busso2008iemocap,
  title={IEMOCAP: Interactive emotional dyadic motion capture database},
  author={Busso, Carlos and Bulut, Murtaza and Lee, Chi-Chun and Kazemzadeh, Abe and Mower, Emily and Kim, Samuel and Chang, Jeannette N and Lee, Sungbok and Narayanan, Shrikanth S},
  journal={Language resources and evaluation},
  volume={42},
  pages={335--359},
  year={2008},
  publisher={Springer}
}

@article{speechteam2024funaudiollm,
  title={FunAudioLLM: Voice Understanding and Generation Foundation Models for Natural Interaction Between Humans and LLMs},
  author={SpeechTeam, Tongyi},
  journal={arXiv preprint arXiv:2407.04051},
  year={2024}
}

@article{zeng2024glm,
  title={Glm-4-voice: Towards intelligent and human-like end-to-end spoken chatbot},
  author={Zeng, Aohan and Du, Zhengxiao and Liu, Mingdao and Wang, Kedong and Jiang, Shengmin and Zhao, Lei and Dong, Yuxiao and Tang, Jie},
  journal={arXiv preprint arXiv:2412.02612},
  year={2024}
}

@article{tang2023salmonn,
  title={Salmonn: Towards generic hearing abilities for large language models},
  author={Tang, Changli and Yu, Wenyi and Sun, Guangzhi and Chen, Xianzhao and Tan, Tian and Li, Wei and Lu, Lu and Ma, Zejun and Zhang, Chao},
  journal={arXiv preprint arXiv:2310.13289},
  year={2023}
}

@article{xu2025qwen2,
  title={Qwen2. 5-omni technical report},
  author={Xu, Jin and Guo, Zhifang and He, Jinzheng and Hu, Hangrui and He, Ting and Bai, Shuai and Chen, Keqin and Wang, Jialin and Fan, Yang and Dang, Kai and others},
  journal={arXiv preprint arXiv:2503.20215},
  year={2025}
}

@misc{shao2023characterllmtrainableagentroleplaying,
      title={Character-LLM: A Trainable Agent for Role-Playing}, 
      author={Yunfan Shao and Linyang Li and Junqi Dai and Xipeng Qiu},
      year={2023},
      eprint={2310.10158},
      archivePrefix={arXiv},
      primaryClass={cs.CL},
      url={https://arxiv.org/abs/2310.10158}, 
}

@article{chu2023qwen,
  title={Qwen-audio: Advancing universal audio understanding via unified large-scale audio-language models},
  author={Chu, Yunfei and Xu, Jin and Zhou, Xiaohuan and Yang, Qian and Zhang, Shiliang and Yan, Zhijie and Zhou, Chang and Zhou, Jingren},
  journal={arXiv preprint arXiv:2311.07919},
  year={2023}
}

@article{chu2024qwen2,
  title={Qwen2-audio technical report},
  author={Chu, Yunfei and Xu, Jin and Yang, Qian and Wei, Haojie and Wei, Xipin and Guo, Zhifang and Leng, Yichong and Lv, Yuanjun and He, Jinzheng and Lin, Junyang and others},
  journal={arXiv preprint arXiv:2407.10759},
  year={2024}
}

@article{comanici2025gemini,
  title={Gemini 2.5: Pushing the frontier with advanced reasoning, multimodality, long context, and next generation agentic capabilities},
  author={Comanici, Gheorghe and Bieber, Eric and Schaekermann, Mike and Pasupat, Ice and Sachdeva, Noveen and Dhillon, Inderjit and Blistein, Marcel and Ram, Ori and Zhang, Dan and Rosen, Evan and others},
  journal={arXiv preprint arXiv:2507.06261},
  year={2025}
}

@article{ji2024wavchat,
  title={Wavchat: A survey of spoken dialogue models},
  author={Ji, Shengpeng and Chen, Yifu and Fang, Minghui and Zuo, Jialong and Lu, Jingyu and Wang, Hanting and Jiang, Ziyue and Zhou, Long and Liu, Shujie and Cheng, Xize and others},
  journal={arXiv preprint arXiv:2411.13577},
  year={2024}
}

@article{yang2025qwen3,
  title={Qwen3 technical report},
  author={Yang, An and Li, Anfeng and Yang, Baosong and Zhang, Beichen and Hui, Binyuan and Zheng, Bo and Yu, Bowen and Gao, Chang and Huang, Chengen and Lv, Chenxu and others},
  journal={arXiv preprint arXiv:2505.09388},
  year={2025}
}

@article{guo2025deepseek,
  title={Deepseek-r1: Incentivizing reasoning capability in llms via reinforcement learning},
  author={Guo, Daya and Yang, Dejian and Zhang, Haowei and Song, Junxiao and Zhang, Ruoyu and Xu, Runxin and Zhu, Qihao and Ma, Shirong and Wang, Peiyi and Bi, Xiao and others},
  journal={arXiv preprint arXiv:2501.12948},
  year={2025}
}

@misc{openai2024gpt4technicalreport,
      title={GPT-4 Technical Report}, 
      author={OpenAI and Josh Achiam and Steven Adler and Sandhini Agarwal and Lama Ahmad and Ilge Akkaya and Florencia Leoni Aleman and Diogo Almeida and Janko Altenschmidt and Sam Altman and Shyamal Anadkat and Red Avila and Igor Babuschkin and Suchir Balaji and Valerie Balcom and Paul Baltescu and Haiming Bao and Mohammad Bavarian and Jeff Belgum and Irwan Bello and Jake Berdine and Gabriel Bernadett-Shapiro and Christopher Berner and Lenny Bogdonoff and Oleg Boiko and Madelaine Boyd and Anna-Luisa Brakman and Greg Brockman and Tim Brooks and Miles Brundage and Kevin Button and Trevor Cai and Rosie Campbell and Andrew Cann and Brittany Carey and Chelsea Carlson and Rory Carmichael and Brooke Chan and Che Chang and Fotis Chantzis and Derek Chen and Sully Chen and Ruby Chen and Jason Chen and Mark Chen and Ben Chess and Chester Cho and Casey Chu and Hyung Won Chung and Dave Cummings and Jeremiah Currier and Yunxing Dai and Cory Decareaux and Thomas Degry and Noah Deutsch and Damien Deville and Arka Dhar and David Dohan and Steve Dowling and Sheila Dunning and Adrien Ecoffet and Atty Eleti and Tyna Eloundou and David Farhi and Liam Fedus and Niko Felix and Simón Posada Fishman and Juston Forte and Isabella Fulford and Leo Gao and Elie Georges and Christian Gibson and Vik Goel and Tarun Gogineni and Gabriel Goh and Rapha Gontijo-Lopes and Jonathan Gordon and Morgan Grafstein and Scott Gray and Ryan Greene and Joshua Gross and Shixiang Shane Gu and Yufei Guo and Chris Hallacy and Jesse Han and Jeff Harris and Yuchen He and Mike Heaton and Johannes Heidecke and Chris Hesse and Alan Hickey and Wade Hickey and Peter Hoeschele and Brandon Houghton and Kenny Hsu and Shengli Hu and Xin Hu and Joost Huizinga and Shantanu Jain and Shawn Jain and Joanne Jang and Angela Jiang and Roger Jiang and Haozhun Jin and Denny Jin and Shino Jomoto and Billie Jonn and Heewoo Jun and Tomer Kaftan and Łukasz Kaiser and Ali Kamali and Ingmar Kanitscheider and Nitish Shirish Keskar and Tabarak Khan and Logan Kilpatrick and Jong Wook Kim and Christina Kim and Yongjik Kim and Jan Hendrik Kirchner and Jamie Kiros and Matt Knight and Daniel Kokotajlo and Łukasz Kondraciuk and Andrew Kondrich and Aris Konstantinidis and Kyle Kosic and Gretchen Krueger and Vishal Kuo and Michael Lampe and Ikai Lan and Teddy Lee and Jan Leike and Jade Leung and Daniel Levy and Chak Ming Li and Rachel Lim and Molly Lin and Stephanie Lin and Mateusz Litwin and Theresa Lopez and Ryan Lowe and Patricia Lue and Anna Makanju and Kim Malfacini and Sam Manning and Todor Markov and Yaniv Markovski and Bianca Martin and Katie Mayer and Andrew Mayne and Bob McGrew and Scott Mayer McKinney and Christine McLeavey and Paul McMillan and Jake McNeil and David Medina and Aalok Mehta and Jacob Menick and Luke Metz and Andrey Mishchenko and Pamela Mishkin and Vinnie Monaco and Evan Morikawa and Daniel Mossing and Tong Mu and Mira Murati and Oleg Murk and David Mély and Ashvin Nair and Reiichiro Nakano and Rajeev Nayak and Arvind Neelakantan and Richard Ngo and Hyeonwoo Noh and Long Ouyang and Cullen O'Keefe and Jakub Pachocki and Alex Paino and Joe Palermo and Ashley Pantuliano and Giambattista Parascandolo and Joel Parish and Emy Parparita and Alex Passos and Mikhail Pavlov and Andrew Peng and Adam Perelman and Filipe de Avila Belbute Peres and Michael Petrov and Henrique Ponde de Oliveira Pinto and Michael and Pokorny and Michelle Pokrass and Vitchyr H. Pong and Tolly Powell and Alethea Power and Boris Power and Elizabeth Proehl and Raul Puri and Alec Radford and Jack Rae and Aditya Ramesh and Cameron Raymond and Francis Real and Kendra Rimbach and Carl Ross and Bob Rotsted and Henri Roussez and Nick Ryder and Mario Saltarelli and Ted Sanders and Shibani Santurkar and Girish Sastry and Heather Schmidt and David Schnurr and John Schulman and Daniel Selsam and Kyla Sheppard and Toki Sherbakov and Jessica Shieh and Sarah Shoker and Pranav Shyam and Szymon Sidor and Eric Sigler and Maddie Simens and Jordan Sitkin and Katarina Slama and Ian Sohl and Benjamin Sokolowsky and Yang Song and Natalie Staudacher and Felipe Petroski Such and Natalie Summers and Ilya Sutskever and Jie Tang and Nikolas Tezak and Madeleine B. Thompson and Phil Tillet and Amin Tootoonchian and Elizabeth Tseng and Preston Tuggle and Nick Turley and Jerry Tworek and Juan Felipe Cerón Uribe and Andrea Vallone and Arun Vijayvergiya and Chelsea Voss and Carroll Wainwright and Justin Jay Wang and Alvin Wang and Ben Wang and Jonathan Ward and Jason Wei and CJ Weinmann and Akila Welihinda and Peter Welinder and Jiayi Weng and Lilian Weng and Matt Wiethoff and Dave Willner and Clemens Winter and Samuel Wolrich and Hannah Wong and Lauren Workman and Sherwin Wu and Jeff Wu and Michael Wu and Kai Xiao and Tao Xu and Sarah Yoo and Kevin Yu and Qiming Yuan and Wojciech Zaremba and Rowan Zellers and Chong Zhang and Marvin Zhang and Shengjia Zhao and Tianhao Zheng and Juntang Zhuang and William Zhuk and Barret Zoph},
      year={2024},
      eprint={2303.08774},
      archivePrefix={arXiv},
      primaryClass={cs.CL},
      url={https://arxiv.org/abs/2303.08774}, 
}

@misc{poria2019meldmultimodalmultipartydataset,
      title={MELD: A Multimodal Multi-Party Dataset for Emotion Recognition in Conversations}, 
      author={Soujanya Poria and Devamanyu Hazarika and Navonil Majumder and Gautam Naik and Erik Cambria and Rada Mihalcea},
      year={2019},
      eprint={1810.02508},
      archivePrefix={arXiv},
      primaryClass={cs.CL},
      url={https://arxiv.org/abs/1810.02508}, 
}

@article{xu2025qwen3,
  title={Qwen3-omni technical report},
  author={Xu, Jin and Guo, Zhifang and Hu, Hangrui and Chu, Yunfei and Wang, Xiong and He, Jinzheng and Wang, Yuxuan and Shi, Xian and He, Ting and Zhu, Xinfa and others},
  journal={arXiv preprint arXiv:2509.17765},
  year={2025}
}

@misc{luo2026surveylargeaudiolanguage,
      title={A Survey of Large Audio Language Models: Generalization, Trustworthiness, and Outlook}, 
      author={Kaiwen Luo and Zhenhong Zhou and Leo Wang and Liang Lin and Yang Xiao and Tianyu Shao and Yuanhe Zhang and Yuxuan Li and Miao Yu and Kailin Lyu and Jiaming Zhang and Dongrui Liu and Li Sun and Yueming Wu and Kai Li and Ting Dang and Xiaojun Jia and Rohan Kumar Das and Xinfeng Li and Siyuan Liang and Qiufeng Wang and Xingjun Ma and Jing Chen and Kun Wang and Junhao Dong and Deqing Zou and Yu Cheng and Xia Hu and Zhigang Zeng and Sen Su and Yang Liu and Yu-Gang Jiang and Philip S. Yu and Yew-Soon Ong},
      year={2026},
      eprint={2605.20266},
      archivePrefix={arXiv},
      primaryClass={cs.SD},
      url={https://arxiv.org/abs/2605.20266}, 
}

@misc{guo2026surveyaudioreasoningmultimodal,
      title={A Survey of Audio Reasoning in Multimodal Foundation Models}, 
      author={Zhihan Guo and Wenqian Cui and Guan-Ting Lin and Daxin Tan and Jingyao Li and Qiyong Zheng and Dingdong Wang and Jing Xiong and Han Shi and Jiaya Jia and Irwin King},
      year={2026},
      eprint={2605.21008},
      archivePrefix={arXiv},
      primaryClass={eess.AS},
      url={https://arxiv.org/abs/2605.21008}, 
}

@misc{xie2025audioreasonerimprovingreasoningcapability,
      title={Audio-Reasoner: Improving Reasoning Capability in Large Audio Language Models}, 
      author={Zhifei Xie and Mingbao Lin and Zihang Liu and Pengcheng Wu and Shuicheng Yan and Chunyan Miao},
      year={2025},
      eprint={2503.02318},
      archivePrefix={arXiv},
      primaryClass={cs.SD},
      url={https://arxiv.org/abs/2503.02318}, 
}

@misc{tian2025stepaudior1technicalreport,
      title={Step-Audio-R1 Technical Report}, 
      author={Fei Tian and Xiangyu Tony Zhang and Yuxin Zhang and Haoyang Zhang and Yuxin Li and Daijiao Liu and Yayue Deng and Donghang Wu and Jun Chen and Liang Zhao and Chengyuan Yao and Hexin Liu and Eng Siong Chng and Xuerui Yang and Xiangyu Zhang and Daxin Jiang and Gang Yu},
      year={2025},
      eprint={2511.15848},
      archivePrefix={arXiv},
      primaryClass={cs.AI},
      url={https://arxiv.org/abs/2511.15848}, 
}

@misc{zhong2025omnir1reinforcementlearningomnimodal,
      title={Omni-R1: Reinforcement Learning for Omnimodal Reasoning via Two-System Collaboration}, 
      author={Hao Zhong and Muzhi Zhu and Zongze Du and Zheng Huang and Canyu Zhao and Mingyu Liu and Wen Wang and Hao Chen and Chunhua Shen},
      year={2025},
      eprint={2505.20256},
      archivePrefix={arXiv},
      primaryClass={cs.CV},
      url={https://arxiv.org/abs/2505.20256}, 
}

@misc{fan2025incentivizingconsistenteffectivescalable,
      title={Incentivizing Consistent, Effective and Scalable Reasoning Capability in Audio LLMs via Reasoning Process Rewards}, 
      author={Jiajun Fan and Roger Ren and Jingyuan Li and Rahul Pandey and Prashanth Gurunath Shivakumar and Ivan Bulyko and Ankur Gandhe and Ge Liu and Yile Gu},
      year={2025},
      eprint={2510.20867},
      archivePrefix={arXiv},
      primaryClass={cs.LG},
      url={https://arxiv.org/abs/2510.20867}, 
}

@misc{hinton2015distillingknowledgeneuralnetwork,
      title={Distilling the Knowledge in a Neural Network}, 
      author={Geoffrey Hinton and Oriol Vinyals and Jeff Dean},
      year={2015},
      eprint={1503.02531},
      archivePrefix={arXiv},
      primaryClass={stat.ML},
      url={https://arxiv.org/abs/1503.02531}, 
}

@misc{kim2016sequencelevelknowledgedistillation,
      title={Sequence-Level Knowledge Distillation}, 
      author={Yoon Kim and Alexander M. Rush},
      year={2016},
      eprint={1606.07947},
      archivePrefix={arXiv},
      primaryClass={cs.CL},
      url={https://arxiv.org/abs/1606.07947}, 
}

@misc{sanh2020distilbertdistilledversionbert,
      title={DistilBERT, a distilled version of BERT: smaller, faster, cheaper and lighter}, 
      author={Victor Sanh and Lysandre Debut and Julien Chaumond and Thomas Wolf},
      year={2020},
      eprint={1910.01108},
      archivePrefix={arXiv},
      primaryClass={cs.CL},
      url={https://arxiv.org/abs/1910.01108}, 
}

@misc{tan2023gkdgeneralknowledgedistillation,
      title={GKD: A General Knowledge Distillation Framework for Large-scale Pre-trained Language Model}, 
      author={Shicheng Tan and Weng Lam Tam and Yuanchun Wang and Wenwen Gong and Yang Yang and Hongyin Tang and Keqing He and Jiahao Liu and Jingang Wang and Shu Zhao and Peng Zhang and Jie Tang},
      year={2023},
      eprint={2306.06629},
      archivePrefix={arXiv},
      primaryClass={cs.CL},
      url={https://arxiv.org/abs/2306.06629}, 
}

@misc{gu2026minillmonpolicydistillationlarge,
      title={MiniLLM: On-Policy Distillation of Large Language Models}, 
      author={Yuxian Gu and Li Dong and Furu Wei and Minlie Huang},
      year={2026},
      eprint={2306.08543},
      archivePrefix={arXiv},
      primaryClass={cs.CL},
      url={https://arxiv.org/abs/2306.08543}, 
}

@misc{chang2022distilhubertspeechrepresentationlearning,
      title={DistilHuBERT: Speech Representation Learning by Layer-wise Distillation of Hidden-unit BERT}, 
      author={Heng-Jui Chang and Shu-wen Yang and Hung-yi Lee},
      year={2022},
      eprint={2110.01900},
      archivePrefix={arXiv},
      primaryClass={cs.CL},
      url={https://arxiv.org/abs/2110.01900}, 
}

@misc{baevski2022data2vecgeneralframeworkselfsupervised,
      title={data2vec: A General Framework for Self-supervised Learning in Speech, Vision and Language}, 
      author={Alexei Baevski and Wei-Ning Hsu and Qiantong Xu and Arun Babu and Jiatao Gu and Michael Auli},
      year={2022},
      eprint={2202.03555},
      archivePrefix={arXiv},
      primaryClass={cs.LG},
      url={https://arxiv.org/abs/2202.03555}, 
}

@misc{milbich2020divadiversevisualfeature,
      title={DiVA: Diverse Visual Feature Aggregation for Deep Metric Learning}, 
      author={Timo Milbich and Karsten Roth and Homanga Bharadhwaj and Samarth Sinha and Yoshua Bengio and Björn Ommer and Joseph Paul Cohen},
      year={2020},
      eprint={2004.13458},
      archivePrefix={arXiv},
      primaryClass={cs.CV},
      url={https://arxiv.org/abs/2004.13458}, 
}

@misc{li2026audiocogitodeepaudioreasoning,
      title={Audio-Cogito: Towards Deep Audio Reasoning in Large Audio Language Models}, 
      author={Longhao Li and Hongjie Chen and Zehan Li and Qihan Hu and Jian Kang and Jie Li and Lei Xie and Yongxiang Li},
      year={2026},
      eprint={2604.12527},
      archivePrefix={arXiv},
      primaryClass={eess.AS},
      url={https://arxiv.org/abs/2604.12527}, 
}

@article{lu2025onpolicydistillation,
  author = {Kevin Lu and Thinking Machines Lab},
  title = {On-Policy Distillation},
  journal = {Thinking Machines Lab: Connectionism},
  year = {2025},
  note = {https://thinkingmachines.ai/blog/on-policy-distillation},
  doi = {10.64434/tml.20251026},
}

@misc{song2026surveyonpolicydistillationlarge,
      title={A Survey of On-Policy Distillation for Large Language Models}, 
      author={Mingyang Song and Mao Zheng},
      year={2026},
      eprint={2604.00626},
      archivePrefix={arXiv},
      primaryClass={cs.LG},
      url={https://arxiv.org/abs/2604.00626}, 
}

@misc{cao2026xopdcrossmodalonpolicydistillation,
      title={X-OPD: Cross-Modal On-Policy Distillation for Capability Alignment in Speech LLMs}, 
      author={Di Cao and Dongjie Fu and Hai Yu and Siqi Zheng and Xu Tan and Tao Jin},
      year={2026},
      eprint={2603.24596},
      archivePrefix={arXiv},
      primaryClass={eess.AS},
      url={https://arxiv.org/abs/2603.24596}, 
}

@misc{hu2026cordbridgingaudiotextreasoning,
      title={CORD: Bridging the Audio-Text Reasoning Gap via Weighted On-policy Cross-modal Distillation}, 
      author={Jing Hu and Danxiang Zhu and Xianlong Luo and Dan Zhang and Shuwei He and Yishu Lei and Haitao Zheng and Shikun Feng and Jingzhou He and Yu Sun and Hua Wu and Haifeng Wang},
      year={2026},
      eprint={2601.16547},
      archivePrefix={arXiv},
      primaryClass={cs.SD},
      url={https://arxiv.org/abs/2601.16547}, 
}

@misc{bengio2015scheduledsamplingsequenceprediction,
      title={Scheduled Sampling for Sequence Prediction with Recurrent Neural Networks}, 
      author={Samy Bengio and Oriol Vinyals and Navdeep Jaitly and Noam Shazeer},
      year={2015},
      eprint={1506.03099},
      archivePrefix={arXiv},
      primaryClass={cs.LG},
      url={https://arxiv.org/abs/1506.03099}, 
}

@article{yuan2025naturalreasoning,
  title={Naturalreasoning: Reasoning in the wild with 2.8 m challenging questions},
  author={Yuan, Weizhe and Yu, Jane and Jiang, Song and Padthe, Karthik and Li, Yang and Kulikov, Ilia and Cho, Kyunghyun and Wang, Dong and Tian, Yuandong and Weston, Jason E and others},
  journal={arXiv preprint arXiv:2502.13124},
  year={2025}
}

@inproceedings{
lambert2025tulu,
    title={Tulu 3: Pushing Frontiers in Open Language Model Post-Training},
    author={Nathan Lambert and Jacob Morrison and Valentina Pyatkin and Shengyi Huang and Hamish Ivison and Faeze Brahman and Lester James Validad Miranda and Alisa Liu and Nouha Dziri and Xinxi Lyu and Yuling Gu and Saumya Malik and Victoria Graf and Jena D. Hwang and Jiangjiang Yang and Ronan Le Bras and Oyvind Tafjord and Christopher Wilhelm and Luca Soldaini and Noah A. Smith and Yizhong Wang and Pradeep Dasigi and Hannaneh Hajishirzi},
    booktitle={Second Conference on Language Modeling},
    year={2025},
    url={https://openreview.net/forum?id=i1uGbfHHpH}
}

@inproceedings{sheng2025hybridflow,
  title={Hybridflow: A flexible and efficient rlhf framework},
  author={Sheng, Guangming and Zhang, Chi and Ye, Zilingfeng and Wu, Xibin and Zhang, Wang and Zhang, Ru and Peng, Yanghua and Lin, Haibin and Wu, Chuan},
  booktitle={Proceedings of the Twentieth European Conference on Computer Systems},
  pages={1279--1297},
  year={2025}
}

@article{ma2025mmar,
  title={Mmar: A challenging benchmark for deep reasoning in speech, audio, music, and their mix},
  author={Ma, Ziyang and Ma, Yinghao and Zhu, Yanqiao and Yang, Chen and Chao, Yi-Wen and Xu, Ruiyang and Chen, Wenxi and Chen, Yuanzhe and Chen, Zhuo and Cong, Jian and others},
  journal={arXiv preprint arXiv:2505.13032},
  year={2025}
}

@article{du2025cosyvoice3,
  title={CosyVoice 3: Towards In-the-wild Speech Generation via Scaling-up and Post-training},
  author={Du, Zhihao and Gao, Changfeng and Wang, Yuxuan and Yu, Fan and Zhao, Tianyu and Wang, Hao and Lv, Xiang and Wang, Hui and Shi, Xian and An, Keyu and others},
  journal={arXiv preprint arXiv:2505.17589},
  year={2025}
}

@misc{drossos2019clothoaudiocaptioningdataset,
      title={Clotho: An Audio Captioning Dataset}, 
      author={Konstantinos Drossos and Samuel Lipping and Tuomas Virtanen},
      year={2019},
      eprint={1910.09387},
      archivePrefix={arXiv},
      primaryClass={cs.SD},
      url={https://arxiv.org/abs/1910.09387}, 
}

@inproceedings{kim-etal-2019-audiocaps,
    title = "{A}udio{C}aps: Generating Captions for Audios in The Wild",
    author = "Kim, Chris Dongjoo  and
      Kim, Byeongchang  and
      Lee, Hyunmin  and
      Kim, Gunhee",
    editor = "Burstein, Jill  and
      Doran, Christy  and
      Solorio, Thamar",
    booktitle = "Proceedings of the 2019 Conference of the North {A}merican Chapter of the Association for Computational Linguistics: Human Language Technologies, Volume 1 (Long and Short Papers)",
    month = jun,
    year = "2019",
    address = "Minneapolis, Minnesota",
    publisher = "Association for Computational Linguistics",
    url = "https://aclanthology.org/N19-1011/",
    doi = "10.18653/v1/N19-1011",
    pages = "119--132"
}

@misc{kwon2023efficientmemorymanagementlarge,
      title={Efficient Memory Management for Large Language Model Serving with PagedAttention}, 
      author={Woosuk Kwon and Zhuohan Li and Siyuan Zhuang and Ying Sheng and Lianmin Zheng and Cody Hao Yu and Joseph E. Gonzalez and Hao Zhang and Ion Stoica},
      year={2023},
      eprint={2309.06180},
      archivePrefix={arXiv},
      primaryClass={cs.LG},
      url={https://arxiv.org/abs/2309.06180}, 
}

@misc{wang2026mmsumassivemultitaskspoken,
      title={MMSU: A Massive Multi-task Spoken Language Understanding and Reasoning Benchmark}, 
      author={Dingdong Wang and Junan Li and Jincenzi Wu and Dongchao Yang and Xueyuan Chen and Tianhua Zhang and Helen Meng},
      year={2026},
      eprint={2506.04779},
      archivePrefix={arXiv},
      primaryClass={cs.CL},
      url={https://arxiv.org/abs/2506.04779}, 
}

@misc{sakshi2024mmaumassivemultitaskaudio,
      title={MMAU: A Massive Multi-Task Audio Understanding and Reasoning Benchmark}, 
      author={S Sakshi and Utkarsh Tyagi and Sonal Kumar and Ashish Seth and Ramaneswaran Selvakumar and Oriol Nieto and Ramani Duraiswami and Sreyan Ghosh and Dinesh Manocha},
      year={2024},
      eprint={2410.19168},
      archivePrefix={arXiv},
      primaryClass={eess.AS},
      url={https://arxiv.org/abs/2410.19168}, 
}

@misc{ma2026interspeech2026audioreasoning,
      title={The Interspeech 2026 Audio Reasoning Challenge: Evaluating Reasoning Process Quality for Audio Reasoning Models and Agents}, 
      author={Ziyang Ma and Ruiyang Xu and Yinghao Ma and Chao-Han Huck Yang and Bohan Li and Jaeyeon Kim and Jin Xu and Jinyu Li and Carlos Busso and Kai Yu and Eng Siong Chng and Xie Chen},
      year={2026},
      eprint={2602.14224},
      archivePrefix={arXiv},
      primaryClass={cs.SD},
      url={https://arxiv.org/abs/2602.14224}, 
}

@article{xu2025qwen25omni,
  title   = {Qwen2.5-Omni Technical Report},
  author  = {Xu, Jin and Guo, Zhifang and He, Jinzheng and Hu, Hangrui and He, Ting and Bai, Shuai and Chen, Keqin and Wang, Jialin and Fan, Yang and Dang, Kai and Zhang, Bin and Wang, Xiong and Chu, Yunfei and Lin, Junyang},
  journal = {arXiv preprint arXiv:2503.20215},
  year    = {2025}
}

@inproceedings{fu-etal-2025-pachat,
    title = "{PACHAT}: Persona-Aware Speech Assistant for Multi-party Dialogue",
    author = "Fu, Dongjie  and
      Cheng, Xize  and
      Li, Linjun  and
      Yang, Xiaoda  and
      Yang, Lujia  and
      Jin, Tao",
    editor = "Christodoulopoulos, Christos  and
      Chakraborty, Tanmoy  and
      Rose, Carolyn  and
      Peng, Violet",
    booktitle = "Proceedings of the 2025 Conference on Empirical Methods in Natural Language Processing",
    month = nov,
    year = "2025",
    address = "Suzhou, China",
    publisher = "Association for Computational Linguistics",
    url = "https://aclanthology.org/2025.emnlp-main.1492/",
    doi = "10.18653/v1/2025.emnlp-main.1492",
    pages = "29325--29342",
    ISBN = "979-8-89176-332-6"
}

\newpage
\appendix

\section{Additional Experiment Details}
\label{sec:appendix_train_details}

\paragraph{Training Setup and Hyperparameters.}
We implement our training framework using \texttt{verl}~\citep{sheng2025hybridflow} on top of Megatron-LM, with \texttt{vLLM}~\citep{kwon2023efficientmemorymanagementlarge} serving as the rollout engine. To accommodate both the $30$B student and the $235$B teacher models within a $32\!\times\!$ H20 GPU budget, we employ hybrid parallelism (PP $=2$, TP $=4$, EP $=4$, CP $=1$) alongside memory optimizations, including teacher parameter offloading and full activation recomputation.

During the on-policy optimization stage, we use a batch size of $256$ and a micro-batch size of $4$ per GPU. We bypass Generalized Advantage Estimation (GAE) and instead use a \texttt{naive} advantage estimator, utilizing the per-token signal from our cross-modal log-ratio (Eq.~\ref{eq:objective}). During generation, we sample $K=4$ responses per prompt at a temperature of $1.0$.

\paragraph{Implementation Details.}
To maintain mathematical consistency with Eq.~\ref{eq:objective} when rollouts and updates are asynchronous, we explicitly track and compute the importance ratio $r_{k,t}$ rather than collapsing it to one. Furthermore, to mitigate precision discrepancies in log-probabilities between \texttt{vLLM} inference and the training engine, we exactly recompute all teacher and student log-probabilities within the training engine during the update phase.


\section{Baseline Descriptions}
\label{sec:appendix_baselines}

For completeness, we provide a concise description of every baseline that appears in the main results (Table~\ref{tab:main}) and the CoT-quality evaluation (Table~\ref{tab:mmar}). Baselines are grouped according to the role they play in our comparison.

\subsection{Closed-Source LALMs}
\paragraph{GPT-4o-Audio~\citep{gpt4o}.}
The audio-enabled variant of OpenAI's GPT-4o family, accepting raw speech and audio as a native input modality and producing text (and, in its real-time variant, speech) outputs. It serves as a strong proprietary reference for general audio-language understanding and is queried through the public API in single-turn, non-streaming mode.

\paragraph{Gemini-3-Flash~\citep{comanici2025gemini}.}
A latency-oriented member of Google's Gemini~3 family with native multimodal input over audio, vision, and text. We use it as a representative high-capacity proprietary LALM and evaluate it via its public API under the same single-turn protocol as GPT-4o-Audio.

\subsection{Open-Weight LALMs}
\paragraph{Qwen2-Audio-Instruct~\citep{xu2025qwen2}.}
A widely adopted open-source LALM that couples a Whisper-style audio encoder with a Qwen2 LLM backbone via an audio adapter. Trained predominantly on perception-oriented tasks (ASR, captioning, short-form QA), it is a natural reference for the pre-reasoning generation of LALMs.

\paragraph{GLM-4-Voice~\citep{zeng2024glm}.}
An end-to-end speech-text model built on the GLM-4 backbone that natively tokenizes speech and integrates it with a unified LLM decoder. It is included to contrast a discrete-speech-token paradigm against continuous-audio LALMs.

\paragraph{Qwen2.5-Omni-7B~\citep{xu2025qwen25omni}.}
A 7B omni-modal model that jointly handles audio, vision, and text. It represents the previous generation of the Qwen-Omni line and provides a smaller-scale, instruction-tuned reference under the same model family as our student.

\paragraph{Qwen3-Omni-Instruct~\citep{xu2025qwen3}.}
The instruction-tuned 30B-A3B (MoE) variant of Qwen3-Omni. It is the closest non-thinking sibling of our student and isolates the contribution of \emph{explicit thinking} from the contribution of our distillation procedure.

\paragraph{Qwen3-Omni-Thinking~\citep{xu2025qwen3}.}
The thinking-mode 30B-A3B variant of Qwen3-Omni and the exact initialization of our student model. It is therefore the primary base-model anchor against which all \textsc{X$^3$-OPD} improvements should be read.

\subsection{Reasoning-Specialized LALMs (Section~\ref{sec:mmar}, Table~\ref{tab:mmar})}
For the CoT-quality evaluation we additionally compare against two recent LALMs that are explicitly designed to produce extended chain-of-thought reasoning over audio.

\paragraph{Audio-Reasoner~\citep{xie2025audioreasonerimprovingreasoningcapability}.}
A Qwen2-Audio-based LALM fine-tuned on \textsc{CoTA}, a 1.2M-sample structured chain-of-thought corpus. Audio-Reasoner adopts a fixed four-stage reasoning template---\emph{Planning} $\rightarrow$ \emph{Captioning} $\rightarrow$ \emph{Reasoning} $\rightarrow$ \emph{Summary}---and is trained purely by supervised fine-tuning on offline teacher traces, making it a strong representative of the static offline-CoT paradigm.

\paragraph{Step-Audio-R1~\citep{tian2025stepaudior1technicalreport}.}
A 33B audio reasoning model built on a Qwen2 audio encoder and a Qwen2.5-32B decoder. Step-Audio-R1 explicitly targets the ``textual-surrogate reasoning'' problem via its Modality-Grounded Reasoning Distillation (MGRD) framework, an iterative self-distillation procedure that filters reasoning traces grounded in acoustic cues, followed by Reinforcement Learning with Verified Rewards. It represents the strongest currently public competitor that combines distillation and RL for audio CoT reasoning.

Together, these two baselines bracket the reasoning-LALM design space: Audio-Reasoner relies entirely on offline structured-CoT SFT, whereas Step-Audio-R1 augments distillation with on-policy RL but remains uni-source (self-distilled). Neither one performs \emph{cross-modal} on-policy distillation across a modality-paired symmetric corpus, which is the gap \textsc{X$^3$-OPD} closes.


\section{Ablation Study}
\label{sec:appendix_ablation}

To attribute the empirical gains of \textsc{X$^3$-OPD} to its individual design choices, we perform a systematic ablation along two axes: the composition of the three-tier symmetric corpus (Section~\ref{sec:data}) and the training-recipe choice between offline warm-start and pure on-policy distillation (Section~\ref{sec:opd}). All variants share the identical student initialization, optimizer, schedule, and rollout budget as the full model, so that any observed difference is attributable solely to the ablated component. To best expose tier-specific contributions, we report four representative columns: MMSU-Sem., MMAU-Sound, MMAU-Music, and BIG Bench Audio.

\begin{table*}[t]
\centering
\small
\setlength{\tabcolsep}{8pt}
\renewcommand{\arraystretch}{1.15}
\begin{tabular}{l cccc}
\toprule
\textbf{Variant} & \textbf{MMSU-Sem.} & \textbf{MMAU-Sound} & \textbf{MMAU-Music} & \textbf{BAB} \\
\midrule
\textsc{X$^3$-OPD} (full)                               & \textbf{83.7} & 88.3 & \textbf{72.4} & \textbf{93.6} \\
\midrule
\multicolumn{5}{l}{\emph{Data composition (drop one tier from the symmetric corpus)}} \\
\quad w/o Tier-1 (textual reasoning $\to$ TTS)        & 80.6 \tiny{($-3.1$)} & 87.5 \tiny{($-0.8$)} & 71.6 \tiny{($-0.8$)} & 89.1 \tiny{($-4.5$)} \\
\quad w/o Tier-2 (audio + caption)                    & 82.9 \tiny{($-0.8$)} & 84.2 \tiny{($-4.1$)} & 69.5 \tiny{($-2.9$)} & 90.4 \tiny{($-3.2$)} \\
\quad w/o Tier-3 (prosody-aware dialogue)             & 80.4 \tiny{($-3.3$)} & 87.6 \tiny{($-0.7$)} & 71.9 \tiny{($-0.5$)} & 91.0 \tiny{($-2.6$)} \\
\midrule
\multicolumn{5}{l}{\emph{Training recipe}} \\
\quad w/o warm-start (pure on-policy)                 & 81.8 \tiny{($-1.9$)} & 86.0 \tiny{($-2.3$)} & 70.7 \tiny{($-1.7$)} & 90.5 \tiny{($-3.1$)} \\
\quad full-data warm-start $\to$ OPD                  & 83.2 \tiny{($-0.5$)} & \textbf{89.1} \tiny{($+0.8$)} & 68.2 \tiny{($-4.2$)} & 92.4 \tiny{($-1.2$)} \\
\bottomrule
\end{tabular}
\caption{Ablation of the three-tier symmetric corpus and the training recipe. Numbers in parentheses indicate the absolute drop relative to the full \textsc{X$^3$-OPD} model.}
\label{tab:ablation}
\end{table*}

\paragraph{Contribution of individual data tiers.}Removing Tier-1 (textual reasoning rendered into speech) causes the largest drop on MMSU-Sem. ($-3.1$) and BAB ($-4.5$), but only marginal regressions on MMAU-Sound/Music. This is the expected signature of Tier-1: it provides the densest source of logical chain-of-thought supervision, and without it the student loses the very capability that the framework is designed to distill. Removing Tier-2 (audio-event reasoning grounded in captions) flips the picture: MMAU-Sound drops by $4.1$ points and MMAU-Music by $2.9$ points, while MMSU-Sem.\ is essentially preserved. This confirms that caption-grounded audio reasoning is responsible for binding the teacher's logical signal to non-linguistic acoustic events. Removing Tier-3 (prosody-aware spoken dialogue) yields a surprisingly steep drop on MMSU-Sem. ($-3.3$), alongside a significant regression on BAB ($-2.6$). This indicates that prosody and conversational context are not merely stylistic wrappers, but rather essential structural cues for deeply parsing the underlying semantics of spoken interactions. Without them, the model's textual-logical reasoning fails to ground properly in real-world conversational audio.

\paragraph{Impact of SFT warm-start duration.}Removing the offline warm-start entirely (\emph{w/o warm-start}) costs $1.9$ points on MMSU-Sem.\ and $3.1$ on BAB. Inspecting the training curves, we observe that pure on-policy training without an SFT phase is markedly more unstable in the early phase: the initial rollouts are far enough from the teacher's distribution that the cross-modal advantage in Eq.~\eqref{eq:objective} is dominated by high-variance corrective terms. The warm-start mitigates this by seating the student inside the teacher's reasoning style before on-policy optimization begins.

\paragraph{Exposure bias in full-corpus warm-starts.}A straightforward alternative is to first SFT the student on the entire three-tier corpus and then run the OPD stage on top (``full-data warm-start $\to$ OPD''). Interestingly, this variant pushes MMAU-Sound even higher than the full model ($+0.8$) but triggers a severe collapse on the held-out MMAU-Music axis ($-4.2$). We interpret this dual effect through two mechanisms. First, as discussed in Section~3, a long offline pass on the full corpus over-fits the student to teacher trajectories under teacher-side inputs $x_t$, entrenching states that the student's acoustic-side rollouts rarely revisit, thereby weakening the corrective signal of the subsequent on-policy phase. Second, extensive SFT on a highly structured, domain-specific corpus inevitably induces catastrophic forgetting of out-of-distribution acoustic domains. While the model over-fits and excels on in-domain sound events, it overwrites the broader perceptual representations (e.g., music) inherited from its pre-training. In future work, expanding the data to a more general distribution may unlock the scaling potential of full-data warm-starts without sacrificing generalizability \citep{lu2025onpolicydistillation}.


\section{Prompt Templates}
\label{sec:appendix_prompts}

This appendix collects the three prompt templates that play a non-trivial role in the \textsc{X$^3$-OPD} pipeline: (i) the teacher-side reasoning prompt used during cross-modal on-policy distillation, (ii) the Tier-1 spoken-style rewriting prompt used to convert textual reasoning corpora into TTS-compatible utterances, and (iii) the Tier-2 caption-reconstruction prompt used to upgrade existing audio-caption datasets into reasoning-ready acoustic descriptions. Table~\ref{tab:prompts} summarizes where each prompt is used in the pipeline, and the verbatim templates follow.

\begin{table}[h]
\centering
\small
\renewcommand{\arraystretch}{1.15}
\resizebox{\columnwidth}{!}{%
\begin{tabular}{l l l}
\toprule
\textbf{Prompt} & \textbf{Stage} & \textbf{Consumer} \\
\midrule
P1: Teacher CoT       & OPD scoring (Sec.~\ref{sec:opd})          & Qwen3-235B-A22B-Thinking \\
P2: Tier-1 TTS rewrite & Tier-1 data construction (Sec.~\ref{sec:data}) & Gemini-3-Flash-Preview \\
P3: Caption reconstruction & Tier-2 data construction (Sec.~\ref{sec:data}) & Qwen3-Omni-Captioner \\
\bottomrule
\end{tabular}%
}
\caption{Summary of prompt templates used in the \textsc{X$^3$-OPD} pipeline.}
\label{tab:prompts}
\end{table}

\paragraph{P1: Teacher chain-of-thought prompt (OPD stage).}
During on-policy distillation, every student rollout $y^{(k)}$ is re-scored token-by-token by the frozen text teacher under the matched text input $x_t$, the question $q$, and the gold answer $a^\star$. The prompt is designed so that the teacher's resulting per-token log-probabilities form a high-confidence, answer-anchored reference distribution against which the student trajectory is contrasted (Eq.~\ref{eq:objective}). 

\begin{figure*}[htbp]
\begin{promptbox}[Prompt 1: Teacher Chain-of-Thought]
You are an expert reasoning teacher. You will be given a description of an acoustic scene or utterance (TEXT), a question about it (QUESTION), and the verified correct answer (ANSWER).

Your task is to produce a faithful, step-by-step chain-of-thought that leads from the TEXT and QUESTION to the ANSWER.

Strict requirements:
1. Reason \emph{only} from evidence available in TEXT. Do not invent acoustic content that is not stated.
2. Cite the specific cue (event, prosody, lexical content, speaker turn, temporal order) you rely on at each step.
3. The reasoning chain must terminate in the exact ANSWER provided; do not propose an alternative.
4. Keep the chain concise and logically tight; avoid filler, restatement, or hedging.

Output format:
<think> ... step-by-step reasoning ... </think>
<answer> ... ANSWER ... </answer>

TEXT: \{x\_t\}
QUESTION: \{q\}
ANSWER: \{a$^\star$\}
\end{promptbox}
\end{figure*}

\paragraph{P2: Tier-1 spoken-style rewriting prompt.}
Tier-1 takes prompts from Tulu~3 and NaturalReasoning, which are written in dense textual style with punctuation, formulae, URLs, and markdown that do not survive TTS rendering. The following prompt is used with Gemini-3-Flash-Preview to normalize each instance into a form that the downstream TTS engine can voice naturally, and that the closed-loop ASR re-checker can verify.

\begin{figure*}[htbp]
\begin{promptbox}[Prompt 2: Spoken-Style Rewriting]
You are an expert in text normalization for TTS (Text-to-Speech) systems. Your task is to process the user's input text according to the following strict rules:

1. Language Check: If the primary language of the text is neither Chinese (ZH) nor English (EN), output exactly the digit \texttt{0} and nothing else.
2. TTS Normalization: Identify special characters, symbols, numbers, abbreviations, URLs, or markdown formatting and convert them into their natural spoken equivalents in the corresponding language. Remove pure visual symbols that have no spoken meaning.
3. Keep it Original: Do NOT rewrite, summarize, or translate regular words. If the original text is already fully suitable for TTS reading, output the original text exactly as it is.
4. Output Format: Output ONLY the final processed text or \texttt{0}. Do not include any explanations, quotes, or markdown formatting.
\end{promptbox}
\end{figure*}

\paragraph{P3: Tier-2 caption reconstruction prompt.}
The original AudioCaps and Clotho captions are written for short captioning evaluation and are typically a single under-specified sentence. To turn them into reasoning-ready descriptions, every audio clip is re-captioned by Qwen3-Omni-30B-A3B-Captioner under the following structured prompt. The reconstructed caption is then used as the teacher-side input $x_t$ for Tier-2.

\begin{figure*}[htbp]
\begin{promptbox}[Prompt 3: Caption Reconstruction]
You are an expert audio annotator. You will receive a short audio clip together with its existing reference caption.

Your task is to produce a single, well-formed acoustic description that is precise enough to support multi-step reasoning over the clip.

The description must explicitly cover, in this order, whenever the relevant evidence is present in the audio:
1. \textbf{Scene}: the overall acoustic environment (indoor / outdoor / studio / vehicle / crowd / nature, etc.).
2. \textbf{Sound events}: every salient event, with a concrete noun-phrase label (e.g. ``glass breaking'', ``female laughter'', ``electric guitar riff'').
3. \textbf{Sources}: the inferred source of each event (human / animal / mechanical / musical / environmental) and, if applicable, the speaker's gender or age band.
4. \textbf{Temporal structure}: the temporal order and relative duration of events (sequential / overlapping / continuous / one-shot).
5. \textbf{Salient acoustic attributes}: pitch, loudness, timbre, tempo, or prosodic cues that are informative for reasoning.

Strict requirements:
a) Describe only what is audibly present. Do not infer external context, intent, or narrative.
b) Do not copy the reference caption verbatim; use it only as a sanity check.
c) If an attribute cannot be determined from the audio, omit it rather than guess.
d) Output one paragraph of plain text, no bullet points, no markdown.

REFERENCE CAPTION: \{caption\_ref\}
AUDIO: <audio clip>
\end{promptbox}
\end{figure*}

\end{document}